\documentclass{article}

\usepackage[utf8]{inputenc} 
\usepackage[T1]{fontenc}    
\usepackage{hyperref}       
\usepackage{url}            
\usepackage{booktabs}       
\usepackage{amsfonts}       
\usepackage{nicefrac}       
\usepackage{xfrac}       
\usepackage{microtype}      

\usepackage{pgfplotstable}
\usepackage{pgfplots}
\usepackage{collcell}
\usepackage{xstring}
\usepackage{tabularx}
\usepackage{graphicx}
\usepackage{tabu}
\usepackage{environ}
\usepackage{fp}
\usepackage{upgreek}
\usepackage{amsmath}
\usepackage{amsthm}
\usepackage{enumitem}
\usepackage{bm}
\usepackage{subfiles}
\usepackage{array,multirow}
\usepackage{caption,subcaption}
\usepackage{makecell}
\usepackage{tikz, color}
\usepackage{todonotes}
\usepackage{amsmath,amssymb}
\usetikzlibrary{matrix}
\usetikzlibrary{calc}
\usetikzlibrary{positioning}
\usetikzlibrary{graphs}
\usetikzlibrary{shapes.geometric}

\usepackage{xcolor,colortbl}
\definecolor{matlab-blue}{rgb}{0,0.4470,0.7410}
\definecolor{matlab-orange}{rgb}{0.8500,0.3250,0.0980}
\definecolor{matlab-yellow}{rgb}{0.9290,0.6940,0.1250}

\definecolor{matlab-green}{rgb}{0.4660,0.6740,0.1880}

\definecolor{matlab-red}{rgb}{0.6350,0.0780,0.1840}

\usepackage[accepted]{include/icml2019}

\icmltitlerunning{Towards Inverse Reinforcement Learning for Limit Order Book Dynamics}

\begin{document}

\twocolumn[
\icmltitle{Towards Inverse Reinforcement Learning for Limit Order Book Dynamics}




\begin{icmlauthorlist}
\icmlauthor{Jacobo Roa-Vicens}{jpm,ucl}
\icmlauthor{Cyrine Chtourou}{jpm}
\icmlauthor{Angelos Filos}{ox}
\icmlauthor{Francisco Rul${\cdot}$lan}{ucl}
\icmlauthor{Yarin Gal}{ox}
\icmlauthor{Ricardo Silva}{ucl}
\end{icmlauthorlist}

\icmlaffiliation{jpm}{JPMorgan Chase \& Co., London, United Kingdom}
\icmlaffiliation{ucl}{Department of Statistical Science, University College London, United Kingdom}
\icmlaffiliation{ox}{Department of Computer Science, University of Oxford, United Kingdom}

\icmlcorrespondingauthor{Jacobo Roa-Vicens}{jacobo.roavicens@jpmorgan.com}

\icmlkeywords{Machine Learning, ICML}

\vskip 0.3in
]



\printAffiliationsAndNotice{} 

\begin{abstract}

Multi-agent learning is a promising method to simulate aggregate competitive behaviour in finance. Learning expert agents' reward functions through their external demonstrations is hence particularly relevant for subsequent design of realistic agent-based simulations. Inverse Reinforcement Learning (IRL) aims at acquiring such reward functions through inference, allowing to generalize the resulting policy to states not observed in the past.
This paper investigates whether IRL can infer such rewards from agents within real financial stochastic environments: limit order books (LOB).
We introduce a simple one-level LOB, where the interactions of a number of stochastic agents and an expert trading agent are modelled as a Markov decision process. We consider two cases for the expert's reward: either a simple linear function of state features; or a complex, more realistic non-linear function. 
Given the expert agent's demonstrations, we attempt to discover their strategy by modelling their latent reward function using linear and Gaussian process (GP) regressors from previous literature, and our own approach through Bayesian neural networks (BNN).
While the three methods can learn the linear case, only the GP-based and our proposed BNN methods are able to discover the non-linear reward case. Our BNN IRL algorithm outperforms the other two approaches as the number of samples increases. These results illustrate that complex behaviours, induced by non-linear reward functions amid agent-based stochastic scenarios, can be deduced through inference, encouraging the use of inverse reinforcement learning for opponent-modelling in multi-agent systems.

\end{abstract}

\section{Introduction}

\let\thefootnote\relax\footnotetext{Opinions expressed in this paper are those of the authors, and do not necessarily reflect the view of JP Morgan.}

Limit order books play a central role in the formation
of prices for financial securities in exchanges globally. These systems
centralize limit orders of price and volume to buy or sell certain securities from large numbers of dealers and investors, matching bids and offers in a transparent process.
The dynamics that emerge from this aggregate process \cite{Preis_2006, cont2010orderbook} of competitive behaviour fall naturally within the scope of multi-agent learning, and understanding them is of paramount importance to liquidity providers (known as market
makers) in order to achieve optimal execution of their operations. We focus on
the problem of learning the latent reward function of an expert agent from a set of observations of their external actions in the LOB.

Reinforcement learning (RL) \citep{sutton2018reinforcement} is a formal framework to study sequential decision-making,
particularly relevant for modelling the behaviour of financial agents in environments like the LOB.
In particular, RL allows to model their decision-making process as
agents interacting with a dynamic environment through policies
that seek to maximize their respective cumulative rewards. Inverse reinforcement learning
\cite{Russell98learningagents} is therefore a powerful framework to analyze and model the actions of such agents, aiming at discovering their latent reward functions:
the most "succinct, robust and transferable definition of a task" \cite{ng2000algorithms}.
Once learned, such reward functions can be generalized to unobserved
regions of the state space, an important advantage over other learning methods.

\paragraph{Overview.}
The paper starts at Section 2 by introducing the foundations of IRL, relevant aspects of the maximum causal entropy model used,
and useful properties of Gaussian processes and Bayesian neural networks applied to IRL. A formal description of the
limit order book model follows. In Section 3, we
formulate the one-level LOB as a finite Markov decision process (MDP) and express its dynamics in terms
of a Poisson binomial distribution. 
Finally, we investigate the performance of two well-known IRL methods (maximum entropy IRL and GP-based IRL), and propose and test an additional IRL approach based on Bayesian neural networks. Each IRL method is tested on two versions of the LOB environment, where the reward function of the expert agent may be either a simple linear function of state features, or a more complex and realistic non-linear reward function.

\paragraph{Related Work.}
Agent-based models of financial market microstructure are extensively used \citep{Preis_2006, navarro2017detailed, wang2017spoofing}.
In most setups, mean-field assumptions \citep{lasry2007mean} are made to obtain
closed form expressions for the dynamics of the complex, multi-agent
environment of the exchanges. We make similar assumptions to obtain a
tractable finite MDP model for the one-level limit order book.

Previous attempts have been made to model the evolution of
the behaviour of large populations over discrete state spaces, combining MDPs
with elements of game theory \cite{jiachen2018}, using maximum causal entropy
inverse reinforcement learning. Their focus had been around social media data.
Recently, \citet{hendricks2017inferring} used IRL in financial market microstructure
for modelling the behaviour of the different classes of agents involved
in market exchanges (e.g. high-frequency algorithmic market makers, machine traders,
human traders and other investors). We draw inspiration from them, 
and distinguish two types of agents: automatic agents that induce our environment's dynamics, and active expert agents that trade in such environment. We focus on inferring the expert agents' objectives from the point of view of an external observer.

Our simplified MDP model could be seen as a variant of
the multi-agent Blotto environment \citep{borel1921theorie, tukey1949problem, roberson2006colonel, balduzzi2019open}. 
Blotto is a resource distribution game consisting of two opponent armies having each a limited number of soldiers that need to be distributed across multiple areas or battlefields. Each area is won by the army that has the highest number of soldiers. The winner army is the one that has majority over the highest number of battlefields.
This environment is often used to model electoral competition problems where parties have a limited budget and need to reach a maximum number of voters.
In our environment, only two
areas are used (best bid and ask), but
the decisions are conditional to a state, hence the MDP
could be seen as a contextual 2-area Blotto variant.

\paragraph{Contributions.} We propose a multi-agent LOB model which provides the possibility of obtaining transition probabilities in closed form, enabling the use of model-based IRL, without giving up reasonable proximity to real world LOB settings.
The reward functions we propose have a clear financial interpretation, and allow flexible and comparable testing of different IRL methods in a setup that can be scaled to higher dimension versions of the environment. We then provide results for the three IRL methods discussed: maximum entropy IRL, Gaussian process-based IRL, and our IRL approach based on Bayesian neural networks. 
Results show that BNNs are able to recover the target rewards, outperforming comparable methods both in IRL performance and in terms of computational efficiency.

\section{Background} \label{sec:background}

\subsection{Inverse Reinforcement Learning} \label{sub:irl}

We base our IRL experiments on a Markov decision process consisting of a tuple
$\langle \mathcal{S},\mathcal{A},\mathcal{T},r,\gamma, P_{0} \rangle$.
$\mathcal{S}$ represents the state space; $\mathcal{A}$ the set of eligible actions; $\mathcal{T}$ represents the model's transition dynamics, where  $\mathcal{T} (s', a, s) = p(s'|s,a)$ is the transition probability to state $s'$ from state $s$ through action $a$; $r(s, a)$ is the unknown reward function we intend to recover; the discount factor $\gamma$ takes values in [0, 1];  and $P_{0}$ represents the initial state distribution. 

Reinforcement learning aims under its general forward formulation at finding an optimal policy $\pi^{*}$  that maximizes the expected  cumulative reward $\mathbb{E}\big[ \sum_{t=0}^{T} \gamma^t r(s_t)|\pi^*\big]$, where the state-action pairs induced by policy $\pi^*$ and transition dynamics $\mathcal{T}$ are denoted in a sequence or trajectory $\mathbf{x} = \{(s_t, a_t)\}_{t=0}^{T}$.

\begin{figure}[!ht]
\centering
\begin{tikzpicture}[scale=0.85, >=stealth]
\tikzstyle{directed}=[->, thick, shorten >=0.5 pt, shorten <=1 pt]
\tikzstyle{inputBox1}=[rectangle, rounded corners, fill = black!10, inner sep=3pt, minimum size = 6.5mm, thick, draw =black!80, node distance = 20mm, scale=0.75]

\tikzstyle{inputBox2}=[rectangle, rounded corners, fill = black!10, inner sep=3pt, minimum size = 6.5mm, thick, draw =black!80, node distance = 20mm, scale=0.75]

\tikzstyle{action1}=[rectangle, fill = matlab-yellow!50, inner sep=3pt, minimum size = 6.5mm, thick, draw =black!100, node distance = 20mm, text=black!100,scale=0.75]

\tikzstyle{action2}=[rectangle, fill = matlab-yellow!50, inner sep=3pt, minimum width = 6.5mm, minimum height = 6.5 mm, thick, draw =black!80, node distance = 20mm, scale=0.75]

\tikzstyle{param}=[circle, fill = matlab-blue!30, inner sep=3pt, minimum size = 6.5mm, thick, draw =black!80, node distance = 20mm, scale=0.75]

\tikzstyle{qv}=[rectangle,rounded corners, fill =green!30, inner sep=3pt, minimum size = 6.5mm, thick, draw =black!100, text=black!100, node distance = 20mm, scale=0.75]

\tikzstyle{line} = [draw, -latex]

\node[inputBox1] at (-0.8,7.5) (demos){$\mathcal{D}=\{\{(s^n_t, a^n_t)\}_{t=0}^{T}\}_{n=1}^{N}$
};

\node[inputBox2] at (5.5,7.5) (mdpmr) {$MDP \backslash r = \langle \mathcal{S},\mathcal{A},\mathcal{T},\gamma, P_{0} \rangle$};
\node[action1] at (2.5,6.0)  (solve)    {Solve MDP}; 
\node[qv] at (2.5,5.0) (sol) {$\mathcal{Q}(s, a; \hat{r}_{\theta}),\mathcal{V}(s; \hat{r}_{\theta})$};
\node at (6.9, 1.3) {IRL}[scale=0.75];
\node at (4.6, 4.45) {RL}[scale=0.75];
\node at (4.9, 2.5) {\textit{Optimize w.r.t. $\theta$}};
\begin{scope}[opacity=0, blend mode=darken]
\node at(-0.3, 4.0) [rectangle,dashed, right, fill=matlab-red!50,draw opacity=1.0, minimum width = 21.5mm, minimum height = 16.5 mm, draw =black!100, thick, scale =3.0] {};
\node at(0.5, 5.4) [rectangle, dashed, right, fill=blue, draw opacity=1.0, minimum width = 12.5mm, minimum height = 6.5 mm, draw =black!100, thick, scale =3.0] {};
\end{scope}
\node[action2] at (2.5,3.3) (MaxEntObj) {\makecell[l]{Evaluate MaxEnt Objective:
$p(\mathcal{D}| \hat{r}_{\theta})$
}};
\node[param]  at (2.5, 2.0) (reward)     {$\hat{r}_{\theta}$};
\path [line] (demos) [directed] |- (MaxEntObj);
\path [line] (mdpmr) [directed] -| (solve);
\path [line] (reward) [directed] -- ++(+4.0,0) -- ++(0,+1) |- (solve);
\path (solve) edge [directed] (sol);
\path (sol)  edge  [directed] (MaxEntObj);
\path (MaxEntObj)  edge  [directed] (reward);

\end{tikzpicture}
\caption{Flow diagram of maximum entropy-based IRL.}
\label{fig:irl}
\end{figure}
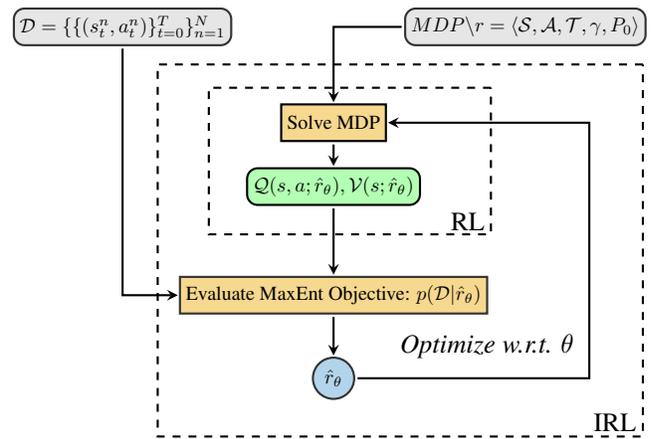

On the other hand, inverse reinforcement learning aims at recovering through inference (see Figure \ref{fig:irl}) an unknown reward function $r(s, a)$, where we assume  $\pi^{*}(a|s)$ to be an optimal policy from where a collection of expert demonstrations $\mathcal{D}=\{\mathbf{x}_{n}\}_{n=1}^{N}$ is drawn. However, one of the main difficulties of the IRL problem is its ill-posed nature, as there could be more than one optimal policy that explains a given set of demonstrations $\mathcal{D}$ \cite{ng1999policyinvariance}. This ambiguity is handled by the maximum entropy framework \citep{ziebart2008maximum}.


\paragraph{Maximum Causal Entropy Model.}

The principle of maximum causal entropy \cite{ziebart2010modeling} assumes that the distribution of trajectories generated by the optimal policy follows $\pi^{*}(a | s) \propto \exp \left\{Q'\left(s_{t}, a_{t}\right)\right\}$ \cite{ziebart2008maximum, haarnoja2017reinforcement}, where $Q'$ is a  is a soft Q-function that incorporates the treatment  of sequentially-revealed information, regularizing the objective of forward reinforcement learning with respect to differential entropy $\mathcal{H}(\cdot)$ as described in \citet{fu2017learning}: $Q'\left(s_{t}, a_{t}\right)=r_{t}(s, a)+\mathbb{E}_{\left(s_{t+1}, \ldots\right) \sim \pi}\left[\sum_{t^{\prime}=t}^{T} \gamma^{t^{\prime}}\left(r\left(s_{t^{\prime}}, a_{t^{\prime}}\right)+\mathcal{H}\left(\pi\left(\cdot | s_{t^{\prime}}\right)\right)\right)\right]$.

The inference problem for an unknown reward function $r(s,a)$ thus boils down to maximizing the likelihood $P(\mathcal{D}|r)$ as in \citet{levine2011nonlinear}:

\vspace{-0.25cm}
\begin{equation}
    exp \sum_{i} \sum_{t} \left[{Q}_{s_{i,t},a_{i,t}}^{'r}-log \sum_{a'} exp  ({Q}_{s_{i,t},{a'}_{i,t}}^{'r} )\right]
    \nonumber
\end{equation}

\subsection{IRL methods considered} \label{sub:irl_methods}

The three inverse reinforcement learning methods that we will test on our LOB model for both linear and exponential expert rewards are: maximum entropy IRL (MaxEnt), Gaussian processes-based IRL (GPIRL), and our implementation through Bayesian neural networks (BNN IRL). These methods are defined as follows:

\paragraph{Maximum Entropy IRL.} Proposed by
\citet{ziebart2010modeling}, this method builds on the maximum causal entropy objective function described above, combined with a linearity assumption on the structure of the reward as a function of the state features:
$ \textbf{r}(s) = \textbf{w}^T\phi(s)$, where $\phi(s): \mathcal{S} \mapsto \mathbb{R}^{m}$ is the $m$-dimensional mapping from the state to the feature vector. The IRL problem then boils down to the inference of such $\textbf{w}$.

\paragraph{Gaussian Process-based IRL.} Maximum causal entropy provides a method to infer values of the reward function on specific points of the state space. This is enough for cases where the MDP is finite and where the observed demonstrations cover all the state space, which are not very common. In this context, it is important to learn a structure of rewards that mirrors as closely as possible the behaviour of the expert agent. 

While many prior IRL methods assume linearity of the reward function, GP-based IRL  \cite{levine2011nonlinear}, expands the function space of possible inferred rewards to non-linear reward structures. Under this model, the reward function $r$ is modeled by a zero-mean GP prior with a kernel or covariance function $k_{\theta}$. If $X_f \in \mathbb{R}^{n,m}$ is the feature matrix defining a finite number $n$ of $m$-dimensional states, and $f$ is the reward function evaluated at $X_f$:  $f=r(X_f)$, then $f|X,\theta \sim \mathcal{N}(0, K_{f,f})$ where  $[K_{f,f}]_{i,j}=k_{\theta}(x_i,x_j)$.

The GPIRL objective \cite{levine2011nonlinear} is then maximized with respect to the finite rewards $f$ and  $\theta$: 
\begin{equation*}
\label{eq:gpirl_obj}
\begin{aligned}
p(f,\theta, \mathcal{D}| X_f) 
&= \Big[\int_{r} p(\mathcal{D}|r) p(r|f, \theta, X_f)dr \Big]  p(f, \theta| X_f)
\end{aligned}
\end{equation*}
Inside the above integral, we can recognize the IRL objective  $p(\mathcal{D}|r)$, the GP posterior $p(r|f, \theta, X_f)$ and the prior of $f$ and $\theta$. To mitigate the intractability of this integral, \citet{levine2011nonlinear} use the Deterministic Training Conditional (DTC) approximation, which reduces the GP posterior to its mean function.

\paragraph{Bayesian Neural Networks applied to IRL.}

A neural network (NN) is a superposition of multiple layers consisting of linear transformations
and non-linear functions. Infinitely wide NNs whose weights are random variables have been proven to converge to a GP \citep{neal1995bayesian, williams1997stochastic} and
hence represent a universal function approximator \citep{cybenko1989approximation}. However, this convergence property is not
applicable to finite NNs, which are the networks used in practice. 
Bayesian neural networks are equivalent networks in the finite case. BNNs have been the focus of multiple studies \citep{neal1995bayesian, mackay1992practical, gal2016dropout} and are known for their useful regularization properties.
Since exact inference is computationally
challenging, variational inference has instead been used to approximate these models \citep{hinton1993keeping, peterson1987mean, graves2011practical, gal2016dropout}.

Given a dataset $\mathcal{Z}=\{(x_{n}, y_{n})\}_{n=1}^{N}$,
we wish to learn the mapping $\{(x_{n}  \rightarrow  y_{n})\}_{n=1}^{N}$ in a robust way. A BNN can be characterized through a prior distribution $p(\mathbf{w})$ placed on its weights, and the likelihood $p(\mathcal{Z}| \mathbf{w})$.
An approximate posterior distribution $q(\mathbf{w})$ can be fit through Bayesian variational inference, maximizing the evidence lower bound (ELBO):

\begin{equation*}
  \mathcal{L}_{q}=\mathbb{E}_{q}[\log p(\mathcal{Z} | \mathbf{w})]-\mathrm{KL}[q(\mathbf{w}) \| p(\mathbf{w})]
\end{equation*}

We parametrize $q(\mathbf{w})$ with $\theta$ and use the mean-field
variational inference approximation \citep{peterson1987mean}, which assumes that the weights of each
layer factorize to independent distributions. We choose a diagonal Gaussian prior distribution $p(\mathbf{w})$
and solve the optimization problem:

\begin{equation*}
    \max _{\theta} \mathcal{L}_{q_{\theta}}
\end{equation*}

In the context of the IRL problem, we leverage the
benefits of BNNs to generalize point estimates provided by maximum
causal entropy to a reward function in a robust and
efficient way. 
Unlike in GPIRL, where the full objective is
maximized at each iteration, BNN IRL consists of two separate steps:
\begin{itemize}
    \item \textbf{Inference step:} we first optimize the IRL objective
    $p(\mathcal{D}|\hat{r})$, obtaining a finite number of point estimates of the reward $\{\hat{r}(s_ n) = \hat{r}_n\}_{n=1}^{N}$ over a finite number of states $\{s_{n} \in \mathcal{S}\}_{n=1}^{N}$
    \item \textbf{Learning step:} secondly, we use these point estimates to train a BNN that learns the mapping between state features $s \in \mathcal{S}$ and their respective inferred reward values $\hat{r}(s) \in \mathbb{R}$.
\end{itemize}

The number of point estimates used is the number of states existing in the expert's demonstrations. This means that we do not use each state in the state space once, but as many times as it exists in the demonstrations $\mathcal{D}$. This is equivalent to tailoring the learning rate of the Bayesian neural network to match the state visitation counts.

\subsection{Limit Order Book} \label{sub:lob}

Our experimental setup builds on limit order books (LOBs):
here we introduce some basic definitions following the
conventions of \citet{gould2013limit} and \citet{zhang2019deeplob}.
Two types of orders exist in an LOB: bids (buy orders) and asks (sell orders).
A bid order for an asset is a quote to buy the underlying asset at or below
a specified price $\mathbf{P}_b(t)$.
Conversely, the ask order is a quote to sell the asset at or above a certain price $\mathbf{P}_a(t)$.
For a certain price, the bid and ask orders have respective sizes (volumes) $\mathbf{V}_b(t)$ and $\mathbf{V}_a(t)$. 
Each level ($l$) of the LOB is represented by one pair ($p^{(l)}_{b}(t), v^{(l)}_{b}(t)$) or ($p^{(l)}_{a}(t), v^{(l)}_{a}(t)$), generally ranked in decreasing order of competitiveness.

\section{Experiments}

\paragraph{Experimental Setup.}

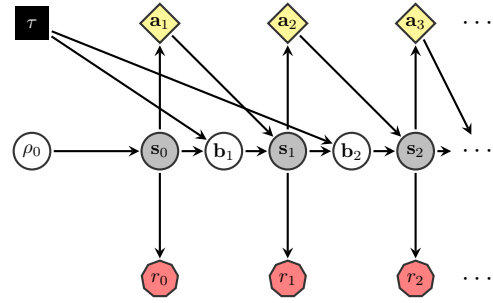
\begin{figure}
\centering
\begin{tikzpicture}[scale=0.85, >=stealth]
\tikzstyle{empty}=[]
\tikzstyle{lat}=[circle, inner sep=1pt, minimum size = 6.5mm, thick, draw =black!80, node distance = 20mm, scale=0.75]
\tikzstyle{obs}=[circle, fill =black!25, inner sep=1pt, minimum size = 6.5mm, thick, draw =black!80, node distance = 20mm, scale=0.75]
\tikzstyle{det}=[rectangle, fill =black, inner sep=1pt, text=white, minimum size = 6mm, draw=none, node distance = 20mm, scale=0.75]
\tikzstyle{dec}=[diamond, fill =yellow!50, inner sep=1pt, minimum size = 6.5mm, thick, draw =black!80, node distance = 20mm, scale=0.75]
\tikzstyle{sca}=[regular polygon,regular polygon sides=9, fill =red!50, inner sep=1pt, minimum size = 6.5mm, thick, draw =black!80, node distance = 20mm, scale=0.75]
\tikzstyle{connect}=[-latex, thick]
\tikzstyle{undir}=[thick]
\tikzstyle{directed}=[->, thick, shorten >=0.5 pt, shorten <=1 pt]

\node[lat] at (1.0,2.0)   (rh0) {$\rho_{0}$};
\node[obs] at (3.0,2.0)   (s0)  {$\mathbf{s}_{0}$};
\node[sca] at (3.0,0.0)   (r0)  {$r_{0}$};
\node[det] at (1.0,4.0)   (tau) {$\mathrm{\tau}$};
\node[lat] at (4.0,2.0)   (b1)  {$\mathbf{b}_{1}$};
\node[dec] at (3.0,4.0)   (a1)  {$\mathbf{a}_{1}$};
\node[obs] at (5.0,2.0)   (s1)  {$\mathbf{s}_{1}$};
\node[sca] at (5.0,0.0)   (r1)  {$r_{1}$};
\node[lat] at (6.0,2.0)   (b2)  {$\mathbf{b}_{2}$};
\node[dec] at (5.0,4.0)   (a2)  {$\mathbf{a}_{2}$};
\node[obs] at (7.0,2.0)   (s2)  {$\mathbf{s}_{2}$};
\node[sca] at (7.0,0.0)   (r2)  {$r_{2}$};
\node[dec] at (7.0,4.0)   (a3)  {$\mathbf{a}_{3}$};
\node      at (8.0,2.0)   (bt)  {$\cdots$};
\node      at (8.0,4.0)   (at)  {$\cdots$};
\node      at (8.0,0.0)   (rt)  {$\cdots$};
\path   
        (rh0) edge [directed] (s0)
        (s0)  edge [directed] (b1)
        (s0)  edge [directed] (r0)
        (tau) edge [directed] (b1)
        (s0)  edge [directed] (a1)
        (a1)  edge [directed] (s1)
        (b1)  edge [directed] (s1)
        (s1)  edge [directed] (b2)
        (s1)  edge [directed] (r1)
        (tau) edge [directed] (b2)
        (s1)  edge [directed] (a2)
        (a2)  edge [directed] (s2)
        (b2)  edge [directed] (s2)
        (s2)  edge [directed] (bt)
        (s2)  edge [directed] (r2)
        (s2)  edge [directed] (a3)
        (a3)  edge [directed] (bt)
        ;
\end{tikzpicture}

\caption{Model of our Markov decision process for a one-level LOB environment.
The expert agent takes actions $\mathbf{a}_{t}$ (diamond yellow nodes), conditioned
on the observed states $\mathbf{s}_{t}$ (shaded circle nodes), receiving reward
$r_{t}$ (polygon red nodes). The trading agents, given the deterministic
temperature parameters $\tau$ (black square node), place orders which comprise
the latent belief state $\mathbf{b}_{t}$. The next state $\mathbf{s}_{t+1}$
is a function of the expert agent's action and the
latent belief states, which the expert agent should learn to infer in order
to optimize cumulative returns.}
\label{fig:dynamics}
\end{figure}

\begin{figure}[!ht]
\centering
\includegraphics[width=\linewidth]{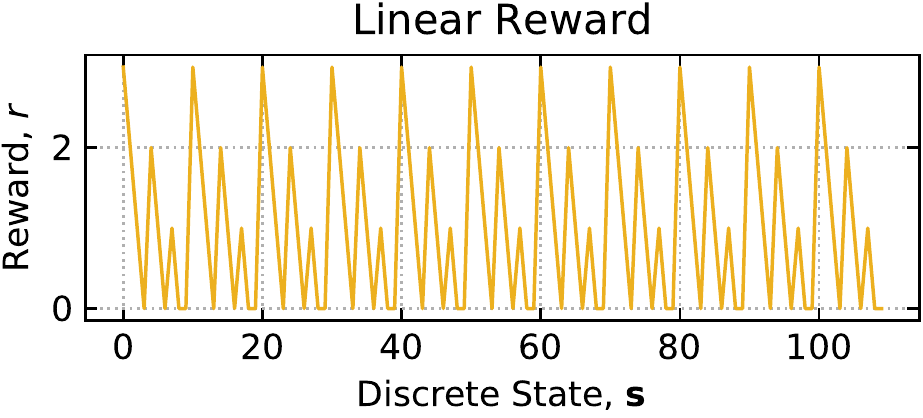}
  
\vspace{0.5em}
  
\includegraphics[width=\linewidth]{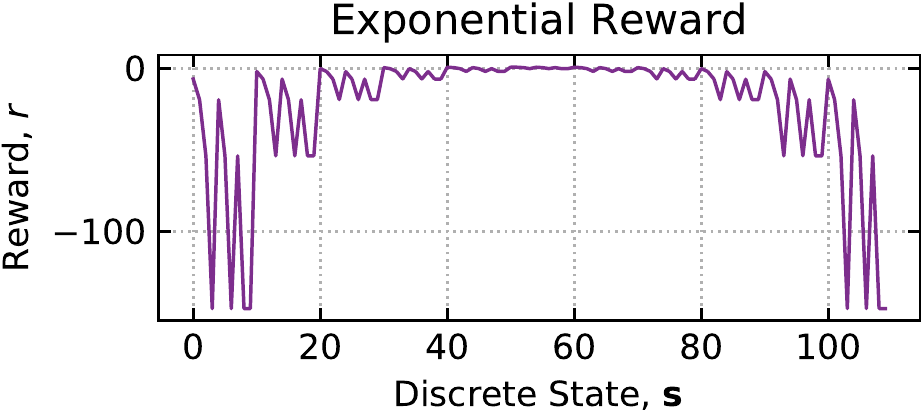}
\caption{Reward functions (i.e. linear and exponential) with respect to the
discrete (enumerated) states, e.g. $\textbf{s}=0$ corresponds to an inventory of $-I_{max}=-5$ and to bid and ask volumes of 0; while $\textbf{s}=1$ corresponds to an inventory of -5, a bid volume of 0 and and ask volume of 1, etc.
}
\label{fig:rewardf}
\end{figure}

\begin{figure}[!ht]
\centering
\begin{subfigure}[l]{0.48\linewidth}
  \includegraphics[width=\linewidth]{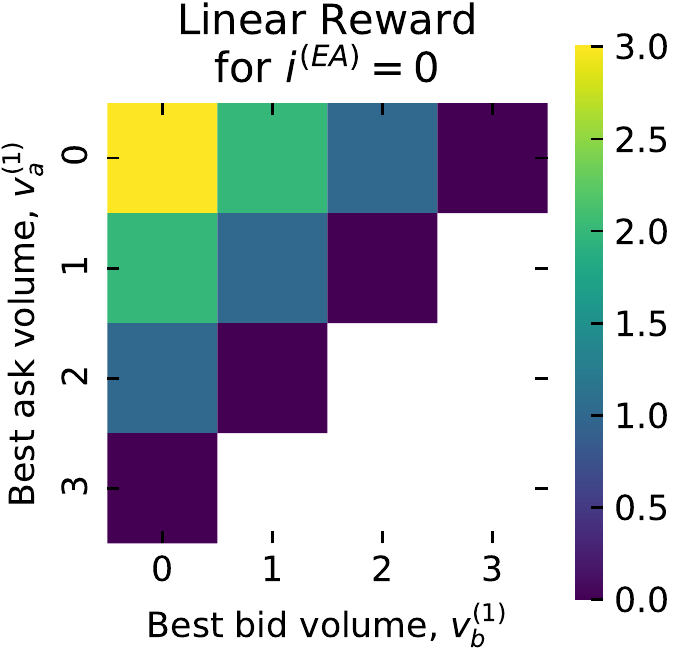}
\end{subfigure}
\begin{subfigure}[l]{0.48\linewidth}
  \includegraphics[width=\linewidth]{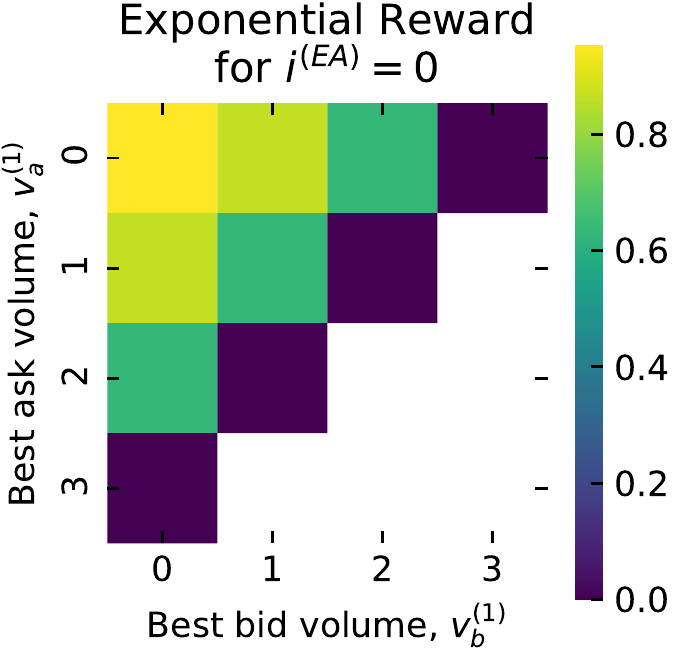}
\end{subfigure}
\caption{Reward functions (linear and exponential) with respect to the
state features: volume of best bid and ask and expert agent's inventory.}
\label{fig:heatmap}
\end{figure}

Our environment setup is a one-level LOB, resulting from the
interaction of $N$ (Markovian) trading agents (TA) and an expert agent (EA trader).
Without loss of generality, the EA is regulated to have up to $I_{\text{max}}$ inventory (amount of securities in their portfolio).
The EA is unaware of the strategies of each
of the other trading agents, but adapts to them through
trial and error, as in reinforcement learning frameworks \citep{sutton2018reinforcement}.
The EA solves the following MDP  $\langle \mathcal{S},\mathcal{A},\mathcal{T},r,\gamma, P_{0} \rangle$:

\begin{itemize}
  \item \textbf{State Space $\mathcal{S}$.}
  The environment state at time-step $t$, $\mathbf{s}_{t}$, is a 3-dimensional vector:
  \begin{equation}
    \mathbf{s}_{t} = \begin{bmatrix} v^{(1)}_{b}(t) & v^{(1)}_{a}(t) & i^{(\text{EA})}(t) \end{bmatrix}^{T} \in \mathcal{S}
  \end{equation}
  where $v^{(1)}_{b}(t)$, $v^{(1)}_{a}(t) \in \mathbb{R}_{+}$ are the volumes available in the one-level LOB through the best bid
  and ask levels, respectively,
  and $i^{(\text{EA})}(t) \in \{-I_{\text{max}}, \ldots, 0, \ldots, +I_{\text{max}} \}$
  is the inventory held by the expert agent at time-step $t$.

  \item \textbf{Action Space $\mathcal{A}$.}
  The expert agent chooses to take an action $\mathbf{a}_{t}$ at time-step $t$, by
  selecting volumes $v^{(EA)}_{b}(t)$ and $v^{(EA)}_{a}(t)$ to match the trading objectives defined by their reward with those available in the LOB through the other
  traders, at the best bid and ask, respectively:
  \begin{equation}
    \mathbf{a}_{t} = \begin{bmatrix} v^{(EA)}_{b}(t) & v^{(EA)}_{a}(t) \end{bmatrix}^{T} \in \mathcal{A}
  \end{equation}
  where $v^{(EA)}_{b}(t) + v^{(EA)}_{a}(t) \leq N$ (see \textit{Transition Dynamics} below).
  The EA is here an active market participant, which actively sells at the best ask and buys at the best bid, while the trading agents on the other side of the LOB only place passive orders.
  \item \textbf{Transition Dynamics $\mathcal{T}$.}
  At each time-step $t$, the $n$-th trader, $n\in \{1, .., N\}$, places a single
  order, $o^{(n)}_t$ at either side of the LOB (best bid, or best
  ask). The traders follow stochastic policies:
  \begin{align}
    \hspace{-0.2cm}\pi^{(n)}( \cdot | \mathbf{s}_{t}; \tau^{(n)} ) &=
      \text{Ber}\left[\frac{e^{\frac{v^{(1)}_{b}(t-1)}{\tau^{(n)}}}}{e^{\frac{v^{(1)}_{b}(t-1)}{\tau^{(n)}}} + e^{\frac{v^{(1)}_{a}(t-1)}{\tau^{(n)}}}} \right] \\
    o^{(n)}_t &\sim \pi^{(n)}( \cdot | \mathbf{s}_{t}; \tau^{(n)} )
  \end{align}
  where $o^{(n)}_t = 1$ corresponds to a trader placing an order at the best bid, and $o^{(n)}_t = 0$ at the best ask. By
  construction, each of the trading
  agents has to place exactly one order, a bid or an ask, at each time-step. Ber($\cdot$) denotes the Bernoulli distribution. The temperature parameters $\tau = (\tau_{1}, \ldots, \tau_{n})$ are independent and unknown to the expert agent.
  
  Hence, the aggregate intermediate bid orders that were generated by the N trading agents, $v^{(TA)}_{b}(t)$ are the sum of $N$ independent
  Bernoulli variables, whose parameter is conditioned on the environment
  state $\mathbf{s}_{t}$ and the idiosyncratic temperature parameters $\tau$.
  
  Therefore, the number of intermediate bids is distributed according to a Poisson binomial distribution \citep{shepp1981entropy},
  whose probability mass function can be expressed in closed form.
  Since each of the trading agents (but not the EA) should place exactly one order, the aggregate intermediate ask orders are $v^{(TA)}_{a}(t) = N - v^{(TA)}_{b}(t)$.
  The aggregate bids and asks of the trading agents form an intermediate state called the belief state $\mathbf{b}_{t} = (v^{(TA)}_{b}(t) , v^{(TA)}_{a}(t))$. \\
  
  Although the EA can only see the last available LOB snapshot contained in $\mathbf{s}_{t}$, their orders ($\mathbf{a}_{t}$) are executed against the intermediate state $\mathbf{b}_{t}$. This could be compared in real world LOBs to a slippage effect. 
  Based on the number of bids and asks finally matched by the EA in the LOB, their inventory is updated as a net long or short position. 
  Considering the above dynamics of the LOB, the
  transition matrix  $\mathcal{T}$, can be calculated exactly. This allows for an exact solution of the MDP. An illustration of
  the dynamics is provided in Figure \ref{fig:dynamics}.

  \item \textbf{Reward Function $r$.}
  The selection of the reward function is crucial, since it
  induces the behaviour of the agent \citep{berger2013statistical}.
  The reward function can be either a function of state $\mathbf{s}_{t}$ and
  action $\mathbf{a}_{t}$; or equivalently, following the dynamics $\mathcal{T}$,
  of the next state $\mathbf{s}_{t+1}$, \citep{sutton2018reinforcement}. We follow the
  latter convention, and use the following two reward functions for the expert agent to test each IRL method considered:
  
  \begin{enumerate}
      \item \textbf{Linear reward}, equal to the EA's hit count:
        \begin{equation}
        r(s) = N - v^{(1)}_{b} - v^{(1)}_{a}
        \end{equation}
    
      This reward is equivalent to the hit count of the EA because there are always exactly N orders in the intermediate state $\mathbf{b}_{t}$, and therefore
      the next state $\mathbf{s}_{t}$ resulting from the execution of the EA's active orders $\mathbf{a}_{t}$ against the TA orders $\mathbf{b}_{t}$, can be effectively seen as the remaining orders, i.e. the number of orders out of a maximum of N that the EA was not able to match.\\

      This reward function does not incentivize directly to sell inventory.
      However, since the episode is terminated when maximum inventory $I_{\text{max}}$
      is exceeded, the market maker is implicitly motivated not to violate this constraint,
      since the simulation will then be terminated and the cumulative reward will be reduced.\\

      \item \textbf{Exponential reward} is a model widely used in economic theory to characterize levels of risk aversion \cite{pratt1964}. We may use it to define a non-linear reward function that depends on both the inventory of the EA and their hit count: 
      \begin{equation}
        r(s) = 1 - e^{- \alpha * (N - v^{(1)}_{b} - v^{(1)}_{a} - \beta  * |i^{(\text{EA})}|) }
      \end{equation}
        where $\alpha, \beta \in \mathbb{R}_{\geq 0}$ are chosen based on the level
        of risk aversion of the agent. 
  \end{enumerate}
  
  Both reward structures are illustrated in Figures \ref{fig:rewardf}- \ref{fig:heatmap}.

\end{itemize}

\paragraph{Results and Discussion.}

Our experiments are performed on synthetic data, where $N=3$ is the
number of trading agents with temperatures $\tau = (0.1, 0.5, 1.0)$, and $I_{\text{max}}=5$.
We also consider an episodic setup, where the simulation is
terminated either when the maximum inventory is violated, or $T=5$ time-steps are completed.
The initial state is uniformly sampled from the non-terminal states in $\mathcal{S}$.

\begin{itemize}
\item \textbf{Performance metric.}
Following previous IRL literature \cite{jin2015inverse, wulfmeier2015maximum} we evaluate the performance of each method through their respective Expected Value Differences (\textbf{EVD}). EVD quantifies the difference between: 
\begin{enumerate}
    \item the expected cumulative reward earned by following the optimal policy $\pi^*$ implied by the true rewards;
    \item the expected cumulative reward earned by following the policy $\hat{\pi}$ implied by the rewards inferred through IRL.
\end{enumerate}
\begin{equation}
    \label{eq: evd} 
    \textbf{EVD} = \mathbb{E}\big[ \sum_{t=0}^{T} \gamma^t r(s_t)|\pi^*\big] - \mathbb{E}\big[ \sum_{t=0}^{T} \gamma^t r(s_t)|\hat{\pi}\big]
\end{equation}

Since the expert's observed behaviour could have been generated by different reward functions, we compare the EVD yielded by inferred rewards per method, rather than directly comparing each inferred reward against the ground truth reward.

\end{itemize}

\begin{figure}[!ht]
  \centering
  \includegraphics[width=0.48\linewidth]{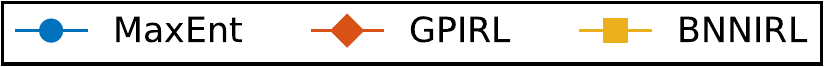}
  
  \vspace{0.5em}
  
  \begin{subfigure}[l]{0.48\linewidth}
    \includegraphics[width=\linewidth]{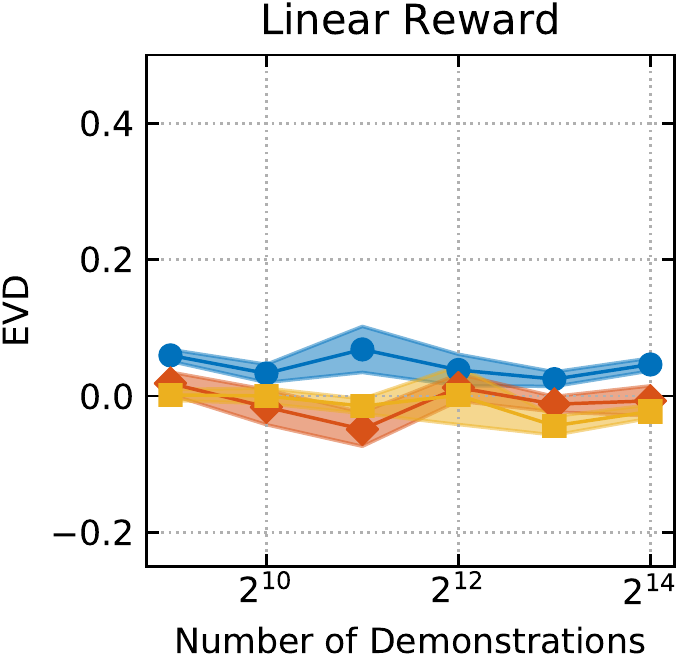}
  \end{subfigure}
  \begin{subfigure}[l]{0.48\linewidth}
    \includegraphics[width=\linewidth]{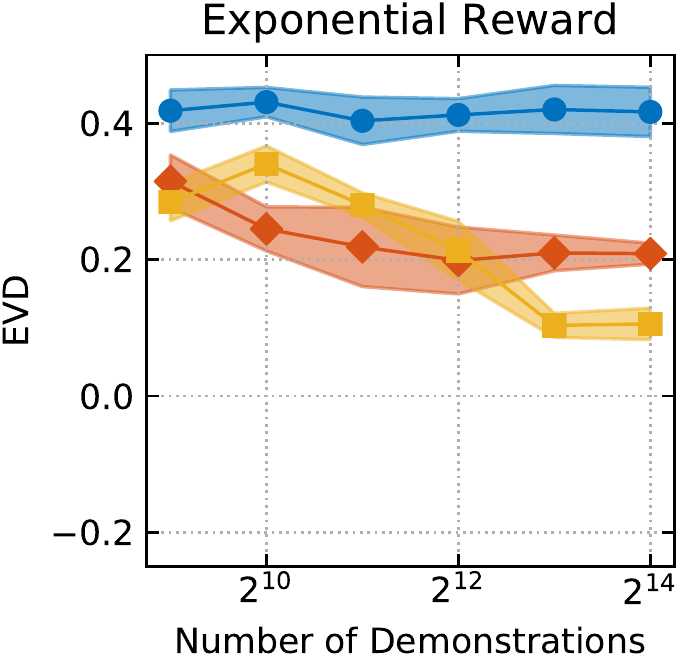}
  \end{subfigure}
  \caption{EVD for both the linear and the exponential reward functions as inferred through MaxEnt, GP and BNN IRL algorithms for increasing numbers of demonstrations. EVDs are all evaluated on 100,000 trajectories each. The standard errors are plotted for 10 independent evaluations.}
  \label{fig:evd}
\end{figure}

We run two versions of our experiments, where the expert
agent has either a linear or an exponential reward function.
Each IRL method is run for 512, 1024, 2048, 4096, 8192
and 16384 demonstrations. The results obtained are presented in Figure
\ref{fig:evd}: as expected, all three IRL methods tested (MaxEnt IRL, GPIRL, BNN-IRL), learn fairly well linear reward functions. However, for an
agent with an exponential reward, GPIRL and BNN-IRL are able to discover the latent function significantly better, with BNN outperforming as the number of demonstrations increases.
In addition to improved EVD, our BNN-IRL experiments provide a significant improvement in computational time as compared to GPIRL, hence enabling potentially more efficient scalability of IRL on LOBs to state spaces of higher dimensions. 

\section{Conclusions}

In this paper we attempt an application of IRL to a stochastic, non-stationary financial environment (the LOB) whose competing agents may follow complex reward functions. While as expected all the methods considered are able to recover linear latent reward functions, only GP-based IRL \cite{levine2011nonlinear} and our implementation through BNNs are able to recover more realistic non-linear expert rewards, thus mitigating most of the challenges imposed by this stochastic multi-agent environment. Moreover, our BNN method outperforms GPIRL for larger numbers of demonstrations, and is less computationally intensive. This may enable future work to study LOBs of higher dimensions, and on increasingly realistic number and complexity of agents involved. 
\bibliography{references}
\bibliographystyle{include/icml2019}

\textbf{Disclaimer}

Opinions and estimates constitute our judgement as of the date of this Material, are for informational purposes only and are subject to change without notice. This Material is not the product of J.P. Morgan’s Research Department and therefore, has not been prepared in accordance with legal requirements to promote the independence of research, including but not limited to, the prohibition on the dealing ahead of the dissemination of investment research. This Material is not intended as research, a recommendation, advice, offer or solicitation for the purchase or sale of any financial product or service, or to be used in any way for evaluating the merits of participating in any transaction. It is not a research report and is not intended as such. Past performance is not indicative of future results. Please consult your own advisors regarding legal, tax, accounting or any other aspects including suitability implications for your particular circumstances. J.P. Morgan disclaims any responsibility or liability whatsoever for the quality, accuracy or completeness of the information herein, and for any reliance on, or use of this material in any way.

Important disclosures at: www.jpmorgan.com/disclosures

\end{document}